\documentclass[10pt,twocolumn,letterpaper]{article}

\usepackage{wacv}
\usepackage{times}
\usepackage{epsfig}
\usepackage{graphicx}
\usepackage{amsmath}
\usepackage{amssymb}

\usepackage{easy-todo}
\usepackage[caption=false]{subfig}
\usepackage[export]{adjustbox}
\def\wacvPaperID{765} % Enter the WACV Paper ID here

\wacvfinalcopy % *** Uncomment this line for the final submission
\ifwacvfinal
\def\assignedStartPage{9876} % *** Enter the assigned starting page number (instead of 9876)
\fi

\ifwacvfinal
\usepackage[breaklinks=true,bookmarks=false]{hyperref}
\else
\usepackage[pagebackref=true,breaklinks=true,colorlinks,bookmarks=false]{hyperref}
\fi

\ifwacvfinal
\else
\fi

\begin{document}

\definecolor{mygreen}{rgb}{0.01, 0.75, 0.24}

%%%%%%%%%%%%%%%%% NEURAL UNITS SYMBOLS DEFINITION %%%%%%%%%%%%%%%%%%

\newcommand{\kernn}[1]{\kern#1pt}
\newcommand{\raisee}[2]{\raisebox{#1}{\ensuremath{#2}}}
\newcommand{\colorr}[2]{\textcolor{#1}{#2}}

%%%%%%%% STNN  %%%%%%
\newcommand{\mtiny}[1]{\mbox{\tiny\ensuremath{#1}}}
\newcommand{\mfootnotesize}[1]{\mbox{\footnotesize\ensuremath{#1}}}
\newcommand{\mscriptsize}[1]{\mbox{\scriptsize\ensuremath{#1}}}
\newcommand{\mlarge}[1]{\mbox{\large\ensuremath{#1}}}
\newcommand{\mlargee}[1]{\mbox{\Large\ensuremath{#1}}}
\newcommand{\mlargeee}[1]{\mbox{\LARGE\ensuremath{#1}}}
\newcommand{\mlargeeee}[1]{\mbox{\huge\ensuremath{#1}}}
\newcommand{\keraise}[3]{\kernn{#2}\raisee{#3pt}{\mtiny{#1}}}
\newcommand{\axstnn}[7]{\begin{array}{#7}\keraise{#1}{#2}{#3}\\\keraise{#4}{#5}{#6}\end{array}}
\newcommand{\bxstnn}[4]{\begin{array}[]{c}\mlargee{#1}\\[#4pt]\keraise{#2}{#3}{0}\end{array}}
\newcommand{\xconv}[5]{\kernn{-4}\axstnn{#4}{0}{0}{#3}{0}{0}{r}\kernn{-12}\bxstnn{\bb{C}}{#5}{8}{-18}
\kernn{-8}\axstnn{#2}{-2.5}{0}{#1}{-2.5}{0}{l}}
\newcommand{\xdense}[5]{\kernn{-6}\axstnn{#4}{0}{0}{#3}{0}{0}{r}\kernn{-14}
\bxstnn{\bb{F}}{#5}{16}{-17}\kernn{-17}\axstnn{#2}{5}{1}{#1}{2}{-1}{l}}
\newcommand{\xpool}[5]{\kernn{-6}\axstnn{#4}{0}{0}{#3}{0}{0}{r}\kernn{-10}
\bxstnn{\bb{P}}{#5}{5}{-15}\kernn{-15}\axstnn{#2}{0}{0}{#1}{0}{0}{l}}
\newcommand{\xinp}[5]{\kernn{-4}\axstnn{#4}{0}{0}{#3}{0}{0}{r}\kernn{-15}
\bxstnn{\cl{I}}{#5}{12}{-17}\kernn{-14}\axstnn{#2}{5}{1}{#1}{2}{-1}{l}}
\newcommand{\xdrop}[5]{\kernn{-2}\axstnn{#4}{0}{0}{#3}{0}{0}{r}\kernn{-15}
\bxstnn{\bb{D}}{#5}{2}{-17}\kernn{-18}\axstnn{#2}{0}{0}{#1}{5}{-1}{l}}
\newcommand{\xmerge}[5]{\kernn{-2}\axstnn{#4}{0}{0}{#3}{0}{0}{r}\kernn{-15}
\bxstnn{\bb{M}}{#5}{1}{-17}\kernn{-9.5}\axstnn{#2}{0}{0}{#1}{0}{-2}{l}}
\newcommand{\xgeneral}[5]{\kernn{-4}\axstnn{#4}{0}{0}{#3}{0}{0}{r}\kernn{-11}
\bxstnn{\bb{Q}}{#5}{0}{-18}\kernn{-9}\axstnn{#2}{-2.5}{0}{#1}{-2.5}{0}{l}}
\newcommand{\xunit}[3]{\kernn{-4}
\overset{#1}{\underset{\raisee{-1.5pt}{\mtiny{#2}}}{\bxstnn{\bb{U}}{#3}{2}{-18}}}}
\newcommand{\xunitdef}[3]{\xunit{#1}{#2}{}
\kernn{-4}\raisee{-2pt}{\ensuremath\longleftarrow\,\boxed{{#3}}}}
\newcommand{\xassign}[2]{{#1}\longleftarrow{#2}}
\newcommand{\xaggreg}[1]{\left\langle{#1}\right\rangle}
%%%%%%%%% OTHERS %%%%%%%%%
\newcommand{\xx}[2]{(#1)(#2)}
\newcommand{\bb}[1]{\mathbb{#1}}
\newcommand{\cl}[1]{\mathcal{#1}}
\newcommand{\tp}[1]{{#1}^{\intercal}}
\newcommand{\tr}[1]{\text{trace}\left[#1\right]}
\newcommand{\inv}[1]{\in\bb{R}^{#1}}
\newcommand{\inm}[2]{\in\bb{R}^{#1\times#2}}
\newcommand{\invc}[1]{\in\bb{C}^{#1}}
\newcommand{\inmc}[2]{\in\bb{C}^{#1\times#2}}
\def\ds{\displaystyle}
\def\ass{\leftarrow}
\def\od#1#2{\nabla_{#2}#1}
\def\tod#1#2{\tp{\nabla}_{#2}{#1}}
\def\cl#1{\mathcal{#1}}

%%%%%%%%%%%%%%%%% END OF NEURAL UNITS SYMBOLS DEFINITION %%%%%%%%%%%%%%%%%%

%%%%%%%%% TITLE
\title{MinkLoc3D: Point Cloud Based Large-Scale Place Recognition}

\author{Jacek Komorowski\\
Warsaw University of Technology\\
Warsaw, Poland\\
{\tt\small jacek.komorowski@pw.edu.pl}
%\and
}

\maketitle
%\thispagestyle{empty}

%%%%%%%%% ABSTRACT
\begin{abstract}
The paper presents a learning-based method for computing a discriminative 3D point cloud descriptor for place recognition purposes.
Existing methods, such as PointNetVLAD, are based on unordered point cloud representation. 
They use PointNet  as the first processing step to extract local features, which are later aggregated into a global descriptor. The PointNet architecture is not well suited to capture local geometric structures. 
%relationships between neighbourhood points 
%Thus, state-of-the-art methods, such as LPD-Net, enhance vanilla PointNet architecture with hand crafted features and attention mechanism using graph convolutional networks. 
%We present an alternative approach, based on sparse voxelized point cloud representation and sparse 3D convolutions. 
%The proposed method has simpler architecture and outperforms state-of-the-art methods by a large margin.
%, is fully learning-based and  produces better results compared to state-of-the-art.
%WHAT ABOUT PERFORMANCE - MEASURE IT COMPARED TO POINTNETVLAD
Thus, state-of-the-art methods enhance vanilla PointNet architecture by adding different mechanism to capture local contextual information, such as graph convolutional networks or using hand-crafted features.
We present an alternative approach, dubbed MinkLoc3D, to compute a discriminative 3D point cloud descriptor, based on a sparse voxelized point cloud representation and sparse 3D convolutions. 
The proposed method has a simple and efficient architecture.
Evaluation on standard benchmarks proves that MinkLoc3D outperforms current state-of-the-art. 
%Comparison with vision-based method proves robustness of MinkLoc3D to difficult environmental conditions.
Our code is publicly available on the project website.~\footnote{\url{https://github.com/jac99/MinkLoc3D}}

\end{abstract}

%%%%%%%%% BODY TEXT

\section{Introduction}

%%%%%%%%%%%%%%%%%%%%%%%%%%%%%%%% WHY THIS RESEARCH EXISTS  AND WHAT IT PRESENTS

\begin{figure}
\centering
\includegraphics[width=1.0\linewidth,trim={0 9.2cm 15cm 0cm},clip]{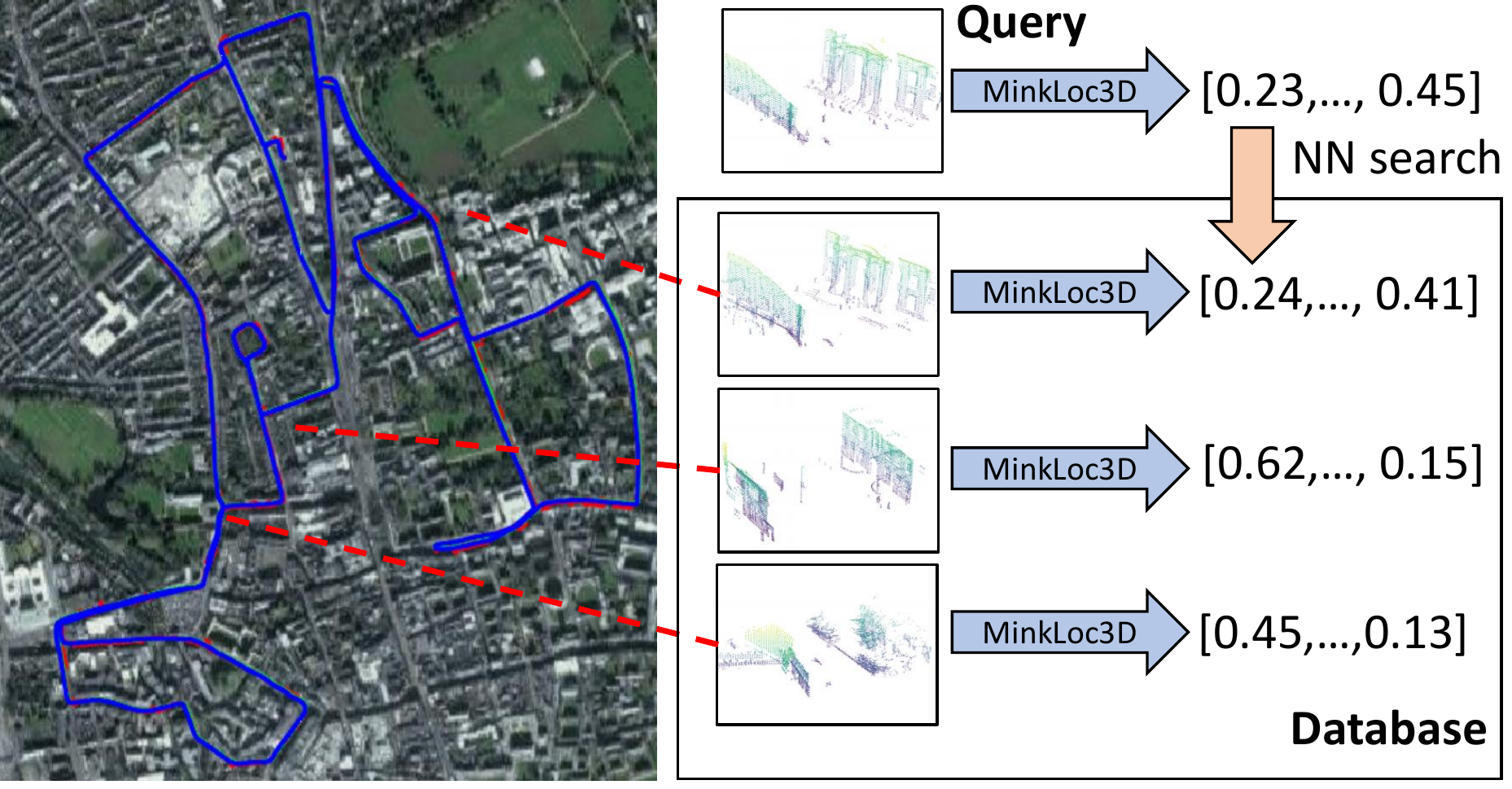}
\caption{Point cloud-based place recognition. MinkLoc3D computes a global descriptor of a query point cloud.
Localization is performed by searching the database for geo-tagged point clouds with closest descriptors.}
\label{fig:teaser}
\end{figure}

Applying deep learning methods to solve 3D computer vision problems 
%based on point cloud representation 
is an area of active development. 
A number of methods for classification~\cite{qi2017pointnet}, %object detection~\cite{wang2019frustum}, 
semantic segmentation~\cite{qi2017pointnet,choy20194d} and local~\cite{choy2019fully} or global~\cite{angelina2018pointnetvlad} features extraction from 3D point clouds was recently proposed.
We focus our attention on finding a discriminative, low-dimensional 3D point cloud descriptor for place recognition purposes.
%Such global descriptors are computed for each processed point cloud and stored in the database. 
Localization is performed by searching the database for geo-tagged point clouds with descriptors closest to the query point cloud descriptor. The idea is illustrated in Fig.~\ref{fig:teaser}.
Place recognition methods are widely used in robotics, autonomous driving~\cite{mcmanus2014shady} and augmented reality~\cite{middelberg2014scalable}.
%In robotics they are used for topological localization, to find loop closure candidates and to reduce drift in SLAM applications.

%\begin{figure*}[tb]

%%%%%%%%%%%%%%%%%%%%%%%%%%%%%%%% WHAT DO WE CHALLENGE HERE

%The main weakness of vision-based place recognition methods is limited robustness for varying image acquisition conditions.
%The appearance of the same scene observed in different atmospheric conditions (e.g. heavy rain or snow, dense fog or poor lighting) or in different times of the day or year can change significantly~\cite{sattler2018benchmarking}.
%In recent years using other modalities to reason about the observed environment gains popularity and became more affordable.
%Lase scanners (LiDAR) are frequently mounted in self-driving cars.
%With introduction of solid state LiDARs they are more and more affordable.
%Active scanning devices, such as LiDARs, provide readings invariant to lighting conditions and to large extent to adverse environmental conditions.

The first learning-based place recognition method operating on 3D point clouds is PointNetVLAD~\cite{angelina2018pointnetvlad}. 
It uses PointNet~\cite{qi2017pointnet} architecture to extract local features and NetVLAD~\cite{arandjelovic2016netvlad} layer to aggregate them into a global descriptor.
%computes a discriminative global descriptor from a raw 3D point cloud, which is used to find and retrieve the most similar point cloud from the database.
%It was shown to perform slightly worse than vision based NetVLAD~\cite{arandjelovic2016netvlad} method in favourable environmental conditions. But it outperformed NetVLAD by a large margin in a challenging day-to-night retrieval scenario.
%But using PointNet architecture to extract local features limits discriminability of the resultant global descriptor.
While PointNet proved to be successful in many applications, it was originally proposed to process point clouds representing single objects, not large and complex scenes. It is not well suited extract informative local features.
%each point is processed in isolation, without taking into account the local neighbourhood. 
%Also intermediate input and feature transform layers do not produce information related to each point neighbourhood. They only regress global transforms that are applied to the entire point cloud.
%Soon, improved 3D place recognition methods such PCAN~\cite{zhang2019pcan} and LPD-Net~\cite{liu2019lpd} were proposed.
To overcome this weakness, latter 3D place recognition methods
%, such as PCAN~\cite{zhang2019pcan} or LPD-Net~\cite{liu2019lpd} 
enhance vanilla PointNet architecture by adding different mechanism to capture local contextual information. 
PCAN~\cite{zhang2019pcan} uses sampling and grouping operation at multiple scales.
%inspired by PointNet++~\cite{qi2017pointnet++} architecture, 
%DAGC~\cite{sun2020dagc} employs graph convolutional networks to aggregate local features of each point at multiple scales.
%All the above mentioned methods use unordered set of points representation and PointNet~\cite{qi2017pointnet} backbone as the first stage of the processing pipeline (this is not true for DAGC).
State-of-the-art LPD-Net~\cite{liu2019lpd} method uses rather complex architecture and combines learning-based and handcrafted local features. 
%First, ten handcrafted features, such as local curvature or point density, are computed for each 3D point. 
3D points enhanced with pre-computed handcrafted features are processed by a PointNet module, fed to a graph neural network to aggregate neighbourhood information and further processed using Point Net architecture. Finally, a global descriptor is computed using NetVLAD~\cite{arandjelovic2016netvlad} layer.
LPD-Net surpasses previous state-of-the-art by a large margin. 
However, at the expense of architectural and computational complexity.
%The step of hand-crafted features extraction is particularly time consuming. For each point 10 different hand-crafted features, each for 10 different neighbourhood sizes, needs to be computed.
%Even using efficient kd-tree implementation to find neighbourhood points, the process takes much longer than rest of the processing pipeline (see Tab.~\ref{jk:tab:resources}).

%%%%%%%%%%%%%%%%%%%%%%%%%%%%%%%% WHAT IS THAT THAT WE PRESENT HERE IN PARTICULAR (method brief and contribution)

Increasing complexity of recent 3D point cloud-based place recognition methods, all based on unordered set of points representation, motivated us to investigate feasibility of using alternate approach and network architecture.
We choose sparse voxelized representation and sparse convolutions, as recently it proved successful in many 3D vision tasks, including local feature extraction~\cite{choy2019fully}, semantic segmentation~\cite{choy20194d} and point cloud registration~\cite{choy2020deep}.

Our method, dubbed MinkLoc3D, has a simple, elegant and effective architecture and outperforms prior state-of-the-art.
MinkLoc3D consists of two parts, local feature extraction network followed by feature aggregation layer.
In order to produce local features with richer semantic content, we adapted Feature Pyramid Network (FPN)~\cite{lin2017feature} architecture. 
The input point cloud is first quantized into a sparse voxelized representation and processed by a local feature extraction network.
%See Feature Extraction block diagram in Fig.~\ref{fig:high_level}.
Different than prior methods we use a simple Generalized-Mean pooling~\cite{lin2017feature} layer, instead of NetVLAD~\cite{arandjelovic2016netvlad} layer, to aggregate local features into a discriminative global point cloud descriptor.

MinkLoc3D achieves state-of-the-art results on standard 3D place recognition benchmarks.
It outperforms PointNetVLAD~\cite{angelina2018pointnetvlad} by a large margin. 
It improves over the current state-of-the-art LPD-Net~\cite{liu2019lpd} despite having simpler architecture and being more computationally effective.
Comparison with vision-based RobotCar Seasons~\cite{sattler2018benchmarking} benchmark proves its robustness to challenging environmental conditions.

Our main contribution is the development of a global point cloud descriptor extraction method based on an alternate point cloud representation and network architecture than prior state-of-the-art.
Our MinkLoc3D method advances state-of-the-art on the popular benchmarks.
It proves the potential of using sparse voxelized representation and sparse convolutions for efficient extraction of discriminative features from 3D point clouds.
We believe our work can spark further improvements in the point cloud-based place recognition field by showing promising development direction. 
%Just like PCAN~\cite{zhang2019pcan} and LPD-Net~\cite{liu2019lpd} improved over PointNetVLAD~\cite{angelina2018pointnetvlad} architecture, our approach can become a base for development of refined methods.

\section{Related work}

\paragraph{Point cloud representation for deep learning.}
Early deep learning methods for 3D point cloud processing use volumetrically discretized representations~\cite{maturana2015voxnet}. 
It's a natural extension of 2D image representation as a grid of pixels and 3D convolutions can be used to effectively process such data. 
However, such representation is very inefficient. The memory requirement grows cubically as spatial resolution increases, making it inappropriate for processing larger point clouds.
%Most of the voxels are empty and processing entire grid of voxels is very inefficient.
%and computationally expensive.

%\cite{su2015multi} proposes multi-view approach, where 2D images of a 3D model are first rendered by multiple virtual cameras placed around the object of interest. Feature maps are computed from these virtual images using 2D convolutional networks, concatenated and fed into the final classification network.

PointNet~\cite{qi2017pointnet} is the first deep learning architecture operating directly on raw 3D point clouds. 
%An input is organized as an unordered set of points, where each point is described by its X, Y, Z coordinates and optional features, such as normal or RGB.
Each point is processed in isolation by multi-layer perceptrons and point features are aggregated using a  symmetric max pooling function.  This makes the architecture independent from input points ordering. 
%PointNet learns a set of functions that select interesting and informative key points from a subset of input points. 
%The advantage of PointNet is it's efficiency, as there's no need to build a costly voxelized representation nor render multiple virtual images. 
%The computational complexity grows linearly with the number of 3D points.
The drawback is that it cannot capture local geometric structures and has limited ability to recognize fine-grained patterns.
To alleviate this problem, PointNet++~\cite{qi2017pointnet++} enhances PointNet  with hierarchical processing.
%First, the input point cloud is partitioned into overlapping local regions.
%using furthest point sampling to choose region centers and finding nearest neighbours of each sampled center. 
%Then, each local region is processed using PointNet to produce a set of local features.
%The above procedure is repeated, at each iteration grouping local features into larger units, until high-level features of the whole point set are computed.
%However, sampling and grouping operation at multiple hierarchy levels is computationally expensive.
%\todo{See wiki article on  a worst case time to construct a kd-tree O(kn log n), where k=3 is dimensionality of the space. Then to find neighbours for n points we need n log n time as well.}

An alternative is to use sparse voxelized representation~\cite{BMVC2015_150}. 
This allows using 3D convolutions to effectively capture local structures and patterns, similarly as 2D convolutions do in 2D images. 
%And sparsity allows compact representation.
However, naive implementations are computationally inefficient.
Recently, an auto-differentiation library for sparse tensors, so called Minkowski Engine \footnote{\url{https://github.com/NVIDIA/MinkowskiEngine}}, was proposed~\cite{choy20194d}.
It efficiently implements sparse convolutions by using coordinate hashing.
%Sparse voxelized representation is very flexible. It allows adapting successful 2D architectures, working on pixel representation, for the 3D world with voxel representation.
%Supporting networks with different architectures, from 3D convolutional networks with ResNet-like architecture to be used for image classification or as a feature extractor, encoder-decoder architectures with residual connections used in semantic segmentation task and variational autoencoders.
Sparse voxelized representation proved successful and yield state-of-the-art results in many 3D vision tasks, such as local feature extraction~\cite{choy2019fully} and semantic segmentation~\cite{choy20194d}.
%and point cloud registration~\cite{choy2020deep}.

%$$\paragraph{Handcrafted point cloud descriptors.}
%\todo{See description in PointNetVLAD paper. They list few methods -- and tell why they are not appropriate for our scenario}

\paragraph{3D point cloud-based place recognition using learned global features.}
PointNetVLAD~\cite{angelina2018pointnetvlad} is the first deep network for large-scale 3D point cloud retrieval. It combines PointNet~\cite{qi2017pointnet} architecture to extract local features and NetVLAD~\cite{arandjelovic2016netvlad} layer to aggregate local features and produce a discriminative global descriptor. 
%The method is evaluated on a subset of Oxford RobotCar Dataset~\cite{RobotCarDatasetIJRR} and in-house datasets prepared by authors.
The main weakness of PointNetVlad is its reliance on PointNet~\cite{qi2017pointnet} for local feature extraction.
PointNet architecture is weak at capturing local geometric structures which adversely impacts discriminability of the resultant global descriptor.
%It processes each point in isolation using a series of multi-layer perceptrons and aggregates them using channel-wise maxpooling operator. This provides invariance to the point order, but does not allow capturing information on local geometric structures.
To overcome this weakness, latter methods enhance PointNetVlad, by adding different mechanism to extract local contextual information.
%All recent methods\cite{angelina2018pointnetvlad,zhang2019pcan,liu2019lpd,sun2020dagc} share the same high-level design. 
%First, the input point cloud is processed using PointNet~\cite{qi2017pointnet} architecture.
%Then, different mechanism is used to compute local features taking into account local geometric properties.
%The final step is NetVLAD~\cite{arandjelovic2016netvlad} layer to aggregate local features into  a discriminative global descriptor.

PCAN~\cite{zhang2019pcan} adds an attention mechanism to predict significance of each point based on a local context.
The input point cloud is first processed using PointNet architecture to compute local features.
Then, sampling and grouping approach inspired by PointNet++~\cite{qi2017pointnet++} is used to extract local contextual information at multiple scales and produce per-point attention map.
Finally, NetVLAD~\cite{arandjelovic2016netvlad} layer aggregates attention-weighted local features into a global descriptor.

%First, local features are computed for each 3D point using PointNet network.
%Then, attention module, with PointNet++\cite{qi2017pointnet++}-like architecture, computes attention map based on local neighbourhood of each point in multiple-scales. Local features are weighted by their significance and fed to the final NetVLAD aggregation layer.

DAGC~\cite{sun2020dagc} combines dynamic graph CNN~\cite{wang2019dynamic} architecture with dual attention mechanism~\cite{fu2019dual} to aggregate local contextual information at multiple scales.
Local features are aggregated using NetVLAD~\cite{arandjelovic2016netvlad} layer to produce a global descriptor.

LPD-Net~\cite{liu2019lpd} relies on handcrafted features and uses graph neural networks to extract local contextual information.
First, ten handcrafted features, such as local curvature or point density, are computed for each point. 
%The features, such as  curvature, local point density and vertical component of a normal vector, describe 1D, 2D and 3D local structures around each point.
Then, 3D points enhanced with handcrafted features are processed using Point Net architecture, fed to a graph neural network to aggregate neighbourhood features and further processed using Point Net-like architecture. Finally, global descriptor is computed using NetVLAD~\cite{arandjelovic2016netvlad} layer.
The method yields state-of-the-art results, surpassing previously proposed solutions by a significant margin.
However, at the expense of architectural complexity and high computational cost. 
%Handcrafted feature extraction is particularly time consuming, as 10 handcrafted features, each at 10 different neighbourhood sizes, are computed for each 3D point. 
%Even using kd-tree to speed-up retrieval of neighbourhood points, the process takes much longer than rest of the processing pipeline (see Tab.~\ref{jk:tab:resources}).

DH3D~\cite{du2020dh3d} is a recent 6DoF relocalization method operating on 3D point clouds.
It unifies global place recognition and local 6DoF pose refinement by inferring local and global 3D descriptors in a single pass through the network.
The local feature extraction module uses Flex Convolution (FlexConv)~\cite{groh2018flex} and Squeeze-and-Excitation (SE)~\cite{hu2018squeeze} blocks to fuse multi-level spatial contextual information and channel-wise feature correlations into local descriptors.
NetVLAD~\cite{arandjelovic2016netvlad} layer aggregates attention-weighted local features into a global point cloud descriptor.
%When evaluated on a point cloud retrieval task using the global descriptor, DH3D falls behind LPD-Net~\cite{liu2019lpd}.%But it must be noted, that the method main purpose is 6DoF relocalization, not the place recognition.

\paragraph{Deep metric learning.} 
Deep metric learning~\cite{lu2017deep} uses deep neural networks to compute a non-linear mapping from a high dimensional data point space to a low-dimensional Euclidean space, known as a representation or embedding space. The learned mapping preserves semantic similarity between objects. 
This technique is widely used in many recognition tasks in computer vision domain, such as pedestrian re-identification~\cite{hermans2017defense} and image retrieval~\cite{lee2008rank}.
%Embeddings of similar data points are closer to each other in a representation space than embeddings of dissimilar objects.
Early deep metric learning methods use a Siamese architecture trained with a contrastive loss~\cite{bromley1994signature}. 
Latter methods propose more complex loss functions, such as triplet~\cite{hermans2017defense} or quadruplet~\cite{chen2017beyond} loss.
Significant attention is put to a selection of an effective sampling scheme to choose informative training samples, so called hard negatives mining~\cite{wu2017sampling}. 
One of the most popular schemes is \emph{batch hard} negative mining proposed in ~\cite{hermans2017defense}, which constructs training triplets by selecting the hardest positive and negative examples within each mini-batch.
%Recent paper~\cite{wang2020cross} identifies a 'slow drift' of embedding features in later epochs of training. 
%Based on this finding authors propose an efficient approach to find informative training examples by using cached embeddings computed from previous mini-batches. This allows selecting hard negatives from much wider pool of examples than a single mini-batch.
In the last few years a number of more sophisticated loss function formulations and sampling schemes
%for deep metric learning 
was proposed~\cite{wu2017sampling,wang2019multi,cakir2019deep}.
However, recent works~\cite{musgrave2020metric,roth2020revisiting} suggest that their advantage over classic contrastive or triplet margin loss is moderate at best. 
%Extensive experimental evaluation shows that their performance is not consistently better than contrastive or triplet losses with properly tuned hyperparameters (values of margin).
Based on these findings we choose triplet margin loss when training our network.

%===========================================================================

\section{ MinkLoc3D: global point cloud descriptor for place recognition}

Our goal is to compute a discriminative and generalizable global descriptor from the input point cloud given as an unordered set of 3D coordinates.
This section describes the proposed architecture and training process of the network computing such descriptor.
%$P = \left\{p_1,\ldots ,p_N \right\} \subset \mathbb{R}^{3}$.
%Such descriptor can be used for visual localization or point cloud retrieval purposes. 
%Descriptors are computed for geo-tagged point clouds and stored in the database. Localization is performed by searching for descriptors closest, in Euclidean distance sense, to the query point cloud descriptor. 
%This allows efficiently retrieving structurally similar point clouds from the database and reason about the localization of the query point cloud.

\subsection{Network architecture}

% Architecture in general

\begin{figure}[tb]
%\centerline{\includegraphics[width=1.0\linewidth]{images/MyMinkNetDiagram.pdf}}
\centerline{\includegraphics[scale=0.8]{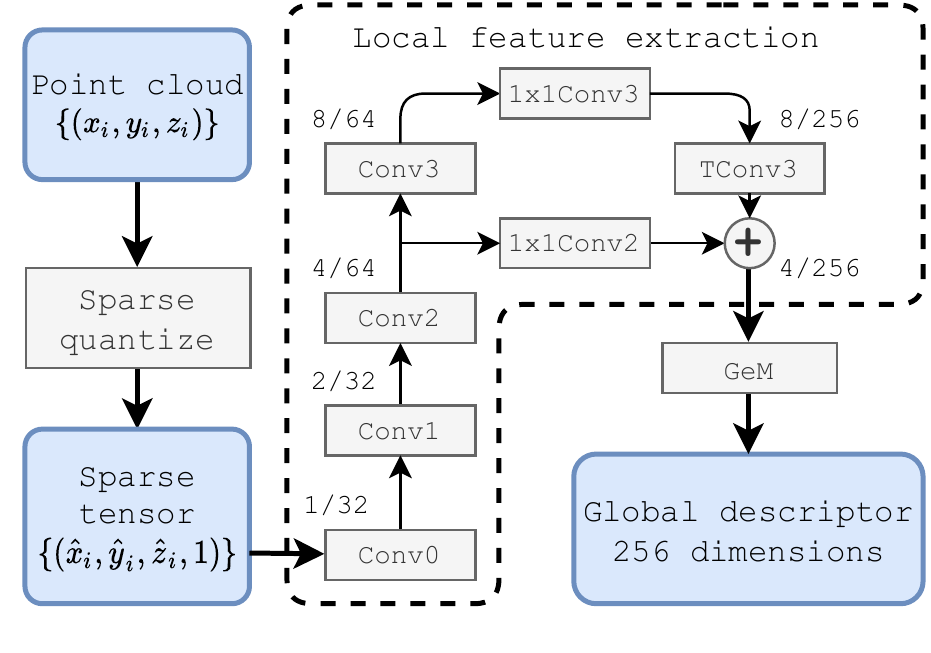}}
\caption{MinkLoc3D architecture. The input point cloud is quantized into a sparse, single channel, 3D tensor. Local features are extracted using a 3D Feature Pyramid Network~\cite{lin2017feature} architecture.
Generalized-mean (GeM)~\cite{radenovic2018fine} pooling produces a global point cloud descriptor. 
Numbers in local feature extraction module (e.g. 1/32) denote a stride and number of channels of a feature map produced by each block.} 
\label{fig:high_level}
\end{figure}

Our network has a very simple architecture shown in Fig.~\ref{fig:high_level}, yet it proved to be more effective and efficient than state-of-the-art methods on standard benchmarks.
It consists of two parts: local feature extraction network and generalized-mean (GeM) pooling~\cite{radenovic2018fine} layer.
Input point cloud $P=\left\{ \left( x_i, y_i, z_i \right) \right\}$, in the form of a set of 3D point coordinates, is first quantized into a single channel sparse tensor $\hat{P}=\left\{ \left( \hat{x}_i, \hat{y}_i, \hat{z}_i, 1 \right) \right\}$.
The values of this single channel are set to one for non-empty voxels.
The sparse tensor if fed to the local feature extraction network, which produces a sparse 3D feature map $\hat{F}=\left\{ \left( \hat{x}_j, \hat{y}_j, \hat{z}_j, f_j^{(1)}, \ldots, f_j^{(c)} \right) \right\}$, where $c$ is a feature dimensionality (256 in our experiments), $\hat{x}_j, \hat{y}_j, \hat{z}_j$ quantized coordinates and $f_j^{(1)}, \ldots, f_j^{(c)}$ features of $j$-th feature map element.
The sparse 3D feature map $\hat{F}$ is pooled using a generalized-mean (GeM) pooling~\cite{radenovic2018fine} layer, which produces a global descriptor vector $\boldsymbol{g}$.
GeM is generalization of a global max pooling and global average pooling operators and is defined as:
$
g^{(k)}
=
\left(
\frac{1}{n}
\sum_{j=1 \ldots  n}
\left(
f_j^{(k)}
\right)
^{p}
\right) 
^
\frac{1}{p}
$, where 
$g^{(k)}$ is $k$-th element of the global descriptor vector $\boldsymbol{g}$, 
$n$ is a size (number of non-zero elements) in the sparse local feature map $\hat{F}$,
$f_j^{(k)}$ is $k$-th feature of the $j$-th local feature map element and $p$ is a learnable pooling parameter.

% And now describe details of each part of the network

The design of the local feature extraction network is inspired by MinkowskiNet~\cite{choy20194d} sparse convolutional network architecture, and Feature Pyramid Network~\cite{lin2017feature} design pattern. 
Bottom-up part part of the network contains four convolutional blocks producing sparse 3D feature maps with decreasing spatial resolution and increasing receptive field.
The top-down part contains a transposed convolution generating an upsampled feature map. 
Upsampled feature map is concatenated with the skipped features from the corresponding layer of the bottom-up pass using a lateral connection.
Such design is intended to produce a feature map with relatively high spatial resolution and large receptive field. Our initial experiments proved its advantage over a simple convolutional architecture without top-down processing. 
%This architecture allows creation of 3D feature map with relatively large spatial resolution and very large receptive fields. The idea of using this design, was to increase the receptive field of each feature map location to allowing capture high-level semantic of the input point cloud.
%\todo{Mention that experimental evaluation and ablation study proves that such design is beneficial over regular convolutional network, without top-down pass.}

\begin{table}[htbp]
\begin{center}
\begin{tabular}{l@{\quad}l}
\hline
Block & Layers
\\[2pt]
\hline
Conv0 & $\xconv{5_{k}1_{s}}{\ 32}{}{}{}$ 
\\[-3pt]
Conv1 &
$\xconv{2_{k}2_{s}}{\ 32}{}{}{}
\Big \langle \;
\xconv{3_{k}1_{s}}{\ 32}{}{}{} \;
\xconv{3_{k}1_{s}}{\ 32}{}{}{}
\Big \rangle
$
\\[-3pt]
Conv2 & 
$\xconv{2_{k}2_{s}}{\ 64}{}{}{}
\Big \langle \;
\xconv{3_{k}1_{s}}{\ 64}{}{}{} \;
\xconv{3_{k}1_{s}}{\ 64}{}{}{}
\Big \rangle
$
\\[-3pt]
Conv3 & 
$\xconv{2_{k}2_{s}}{\ 64}{}{}{}
\Big \langle \;
\xconv{3_{k}1_{s}}{\ 64}{}{}{} \;
\xconv{3_{k}1_{s}}{\ 64}{}{}{}
\Big \rangle
$
\\[-3pt]
1x1Conv2
&
$\xconv{1_{k}1_{s}}{\ 256}{}{}{}$
\\[-3pt]
1x1Conv3
&
$\xconv{1_{k}1_{s}}{\ 256}{}{}{}$
\\[-3pt]
TConv3 &  
$\xconv{2_{k}2_{s}}{\ 256}{t}{}{}$
\\[2pt]
\hline
\end{tabular}
\end{center}
\caption{Details of a local feature extraction part of MinkLoc3D network. All convolutions in bottom-up Conv$0\ldots3$ blocks are followed by batch norm and ReLU non-linearity.
$\langle \ldots \rangle$ denotes a residual block.}
\label{tab:details}
\end{table}

Tab.~\ref{tab:details} shows details of each convolutional block in a local feature extraction network.
We use notation introduced in~\cite{skarbek2019symbolic}, where
$\xconv{a_{k}b_{s}}{\ c}{}{}{}$
denotes a convolution with $c$ kernels of shape $a \times a \times a$ and stride b.
$t$ decorator is used to indicate a transposed convolution.
%$\xconv{}{}{t}{}{}$ 
%, followed by batch normalization and ReLU non-linearity (denoted by $br$), $\xconv{3_{k}}{\ i}{}{}{br}$ denotes convolution with $i$ kernels of 3x3x3 shape and default stride 1.
$\langle \ldots \rangle$ denotes a residual block with a skip connection, defined as $\langle f \rangle \left( x \right) \doteq f \left( x \right) + x$.
%Our network is designed to operate on relatively sparse point clouds.
The first convolutional block (Conv0) has bigger 5x5x5 kernels, in order to aggregate information from a larger neighbourhood.
Bottom-up blocks (Conv1, Conv2 and Conv3) are made of a stride two convolution, which decreases spatial resolution by two, followed by residual block consisting of two convolutional layers with 3x3x3 kernel.
All convolutional layers in bottom-up blocks are followed by batch normalization~\cite{pmlr-v37-ioffe15} layer and ReLU non-linearity.
Two 1x1Conv blocks have the same structure, both contain a single convolutional layer with 1x1x1 kernel.
The aim of these blocks is to unify the number of channels in feature maps produced by bottom-up blocks, before they are merged in the top-down pass through the network.
The top-down part of the network consists of a single transposed convolution layer (TConv3) with 2x2x2 kernel.

%This architecture is very flexible and the network capacity can be easily extended by increasing the network depth or the number of convolutional filters.
%However, experimental evaluation proved that such simple design works surprisingly well, surpassing state-of-the-art methods by a significant margin. 

%It must be noted that, contrary to prior methods such as PointNetVLAD~\cite{angelina2018pointnetvlad} or LPD-Net~\cite{liu2019lpd}, we do not $L_2$-normalize the global descriptor produced by GeM layer. 
%We observed that adding $L_2$-normalization makes the performance of the trained network is very sensitive to the choice of margin hyperparameter in the triplet loss.
%We noticed, that applying $L_2$-normalization produces worse results makes the training very susceptible to the choice of triplet loss margin hyperparameter. 
%Without  $L_2$-normalization the choice of margin hyperparameter is not so critical and the resultant descriptor performs better. More details can be found in Supplementary Material.
%\todo{Mention that we prevent the risk of emnbedding collapse by dynamic batch sizing -- see details in Network trainin section}

\subsection{Network training}

To train our network we use a deep metric learning approach~\cite{lu2017deep} with a triplet margin loss~\cite{hermans2017defense} defined as:
\[
L(a_i,p_i,n_i) = \max \left\{ d(a_i,p_i) - d(a_i,n_i) + m, 0 \right\} ,
\]
where 
\(d(x,y) = || x - y ||_2\) is an Euclidean distance between embeddings $x$ and $y$;  \(a_i, p_i, n_i\) are embeddings of an anchor, a positive and a negative elements in $i$-th training triplet and $m$ is a margin hyperparameter. 
The loss function is minimized using a stochastic gradient descent approach with Adam optimizer. 

Previous methods, such as PointNetVLAD~\cite{angelina2018pointnetvlad} and LPD-Net\cite{liu2019lpd}, use rather inefficient training strategy.
In order to construct informative triplets, for each anchor point cloud they sample 2 positive and 18 negative candidates.
%\footnote{In out training datasets, structurally similar point clouds as defined as point clouds located within 10 meter distance, and dissimilar are further than 50 meters apart. Similarity is indefinite for point clouds between 10 and 50 meters apart.}
Embeddings of all candidate point clouds are calculated and only one hardest positive and negative example is taken to construct a training triplet.
Thus, 21 point clouds need to be processed to construct one triplet.
%In one epoch, an embedding of each training element is computed more than 20 times on average.
%Additionally, in later epochs an offline hard negative mining is added which further increases computational complexity.
%A few times during one epoch, embeddings of all training set elements are computed stored in the memory.
%The pool of negative candidates is constructed by selecting the hardest negative from 18 randomly sampled dissimilar point clouds and few hard negative examples found using all previously cached embeddings.
%This considerably increases time complexity.

We developed an alternative, more efficient, training procedure based on batch hard negative mining approach~\cite{hermans2017defense}.
At the beginning of each epoch we randomly partition a training set into batches.
A batch of size $n$ is constructed by sampling $n/2$ pairs of structurally similar elements. 
%Sampling is done without replacement, that is in one epoch an element is included in only one batch. 
%By the construction process, a batch element $e$ has least one other structurally similar element in the batch. %There may be more, as by chance other examples in the batch can be structurally similar to $e$.
%Because majority of training set elements are structurally dissimilar to $e$, most of the remaining batch elements will be dissimilar to $e$. 
%Thus, for a batch element $e$ usually there are many structurally dissimilar elements, the more the bigger the batch.
After a batch is constructed, we compute two $n \times n$ boolean masks, one indicating structurally similar pairs and the other structurally dissimilar.
We use hash-based indexing to efficiently check if two elements are structurally similar, dissimilar or similarity is indefinite.
Then, the batch is fed to the network to compute embeddings. 
Using similarity and dissimilarity boolean masks and computed embeddings, we mine hardest positive and hardest negative examples and construct informative training triplets.
In our approach, processing one batch consisting of $n$ elements produces $n$ training triplets.
This approach brings down network training time from days to hours.

During experiments we noticed that with larger batch sizes, the training process is prone to collapse, where all embeddings approach the same value.
To overcome this problem, we use a simple yet effective dynamic batch sizing strategy.
The training starts with a small batch size, say 16 examples.  
At the end of each epoch, the average number of active triplets (i.e. triplets producing non-zero loss) per batch is examined.
If the ratio of active triplets to all triplets in a batch falls below the predefined threshold $\Theta$, the batch size is increased by a fixed batch expansion rate $\alpha$.

%In our experiments we use $\Theta = 70\%$ active triplets threshold and $\alpha = 1.4$ batch expansion rate, although the process is relatively insensitive to exact values of these parameters.
%This simple approach works very well in practice. Initial training starts with relative small batch sizes, with moderately hard training examples. As the network learns to recognize these easier examples, which manifests itself by the decreasing number of active triplets, the batch size increases and more difficult training triplets are constructed. 
%The training data complexity gradually increases from relatively easy to more difficult cases.

%\todo{
%We think that our approach is better, as in original:
%- in first epoch there's a small pool to mine negatives
%- in later epochs, when offline triplet mining is added, the examples rapidly become very hard (the hardest %example from all training set) - which adversely impact training process in the presence of noise etc.
%}

To increase variability of the training data and reduce overfitting, we apply on-the-fly data augmentation. 
It includes random jitter with a value drawn from a normal distribution $\mathcal{N} \left(\mu=0, \sigma=0.001 \right)$; random translation by a value sampled from $0 \ldots 0.01$ range; and removal of randomly chosen points, where the percentage of points to remove is uniformly sampled from $0 \ldots 10\%$ range.
We also adapted a random erasing augmentation~\cite{zhong2017random} and randomly remove all points within a fronto-parallel cuboid with a random size and position.

\section{Experimental results}

In this section we describe the datasets and evaluation methodology, compare our method to the state-of-the-art and conduct ablation study. We also compare our method with image-based visual localization method on the standard visual localization benchmark.

\subsection{Datasets and evaluation methodology}

The network is trained and evaluated using a modified Oxford RobotCar dataset and three in-house datasets: University Sector (U.S.), Residential Area (R.A.) , Business District (B.D.) introduced in~\cite{angelina2018pointnetvlad}. 
The datasets are created using a LiDAR sensor mounted on the car travelling through these four regions at different times of day and year.
Oxford RobotCar dataset is build using SICK LMS-151 2D LiDAR scans and in-house dataset using Velodyne HDL-64 3D LiDAR. 
%These are standard datasets used to report performance of recent point cloud-based place recognition methods such as PointNetVLAD~\cite{angelina2018pointnetvlad}, PCAN\cite{zhang2019pcan}, DAGC~\cite{sun2020dagc} and LPD-Net~\cite{liu2019lpd}.

All point clouds are preprocessed with the ground planes removed and downsampled to 4096 points. The point coordinates are shifted and rescaled to be zero mean and inside the $\left[-1, 1\right]$ range. See Fig.~\ref{fig:search_results} for exemplary data items.
Training tuples are generated using ground truth UTM coordinates. Structurally similar point clouds are at most 10m apart. Dissimilar point clouds are at least 50m apart.
For point clouds between 10 and 50m apart similarity is indefinite. Each dataset is split into disjoint training and test subsets. For more information please refer to~\cite{angelina2018pointnetvlad}.

Same as in previous works, we evaluate our network in two scenarios.
In \emph{baseline} scenario, the network is trained using the training subset of Oxford dataset and evaluated on test splits of Oxford and in-house datasets.
%This allows assessing generalization capability, as Oxford and in-house datasets were gathered using different types of LiDAR.
In \emph{refined} scenario, the network is trained on the training subset of Oxford and in-house datasets; and evaluated on test splits of Oxford and in-house datasets.
The number of training and test elements used in each scenario is shown in Tab.~\ref{jk:tab:datasets}.

\begin{table}[htbp]
\begin{center}
\begin{tabular}{l@{\quad}r@{\quad}r@{\quad}r@{\quad}r}
\hline
& \multicolumn{2}{c}{Baseline Dataset} 
& \multicolumn{2}{c}{Refined Dataset} 
\\
& \begin{tabular}{@{}c@{}}Training \end{tabular}
& \begin{tabular}{@{}c@{}}Test \end{tabular}
& \begin{tabular}{@{}c@{}}Training \end{tabular}
& \begin{tabular}{@{}c@{}}Test \end{tabular}
\\
[2pt]
\hline
Oxford & 21.7k & 3.0k & 21.7k & 3.0k \\
In-house
& \begin{tabular}{@{}c@{}} - \end{tabular} & 4.5k & 6.7k & 1.7k \\
[2pt]
\hline
\end{tabular}
\end{center}
\caption{Number of elements in datasets used in \emph{baseline} and \emph{refined} evaluation scenarios.}
\label{jk:tab:datasets}
\end{table}

\paragraph{Evaluation metrics} 
%We follow the standard place recognition evaluation procedure~\cite{arandjelovic2016netvlad,angelina2018pointnetvlad,liu2019lpd}.
We follow the same evaluation protocol as in~\cite{angelina2018pointnetvlad,liu2019lpd}.
A point cloud from a testing dataset is taken as a query and point clouds from different traversals that cover the same region form the database. 
The query point cloud is successfully localized if at least one of the top $N$ retrieved database clouds is within $d=25$ meters from
the ground truth position of the query.
\emph{Recall@N} is defined as the percentage of correctly localized queries.
As in~\cite{angelina2018pointnetvlad} we report Average Recall@1 (AR@1) and Average Recall@1\% (AR@1\%) metrics.

\paragraph{Implementation details.} 

In all experiments we quantize 3D point coordinates with 0.01 quantization step. As point coordinates in the Baseline and Refined datasets are normalized to be in $\left[-1, 1\right]$ range, this produces up to 200 voxels in each spatial direction.
Other parameters of the training process are listed in~Tab.\ref{jk:tab:params}.
Initial learning rate is divided by 10 at the epoch given in LR scheduler steps row. The Refined Dataset is larger and more diverse than Baseline Dataset, hence in refined scenario the network is trained twice as long.
The dimensionality of the resultant global descriptor is set to 256 , same as in prior methods.

\begin{table}[htbp]
\begin{center}
\begin{tabular}{l@{\quad}l@{\quad}l}
\hline
& \begin{tabular}{@{}c@{}}Baseline \end{tabular}
& \begin{tabular}{@{}c@{}}Refined\end{tabular}
\\
[2pt]
\hline
Initial batch size & 32 & 16 \\
% Changed 11-08
Batch size limit & 256 &  256 \\
Batch expansion threshold $(\Theta)$ & 0.7 & 0.7 \\
Batch expansion rate $(\alpha)$ & 1.4 &  1.4 \\
Number of epochs & 40 &  80 \\
Initial learning rate & 1e-3 & 1e-3 \\
LR scheduler steps & 30 &  60 \\
% Added 11-08
$L_2$ weight decay & 1e-3 &  1e-3 \\
Triplet loss margin $(m)$ & 0.2 & 0.2  \\
[2pt]
\hline
\end{tabular}
\end{center}
\caption{Parameters of the training process in \emph{baseline} and \emph{refined} evaluation scenarios.}
\label{jk:tab:params}
\end{table}

All experiments are performed on a server with a single nVidia RTX 2080Ti GPU, 12 core AMD Ryzen Threadripper 1920X processor, 64 GB of RAM and SSD hard drive. We use PyTorch 1.5~\cite{NEURIPS2019_9015} deep learning framework, MinkowskiEngine 0.4.3~\cite{choy20194d} auto-differentiation library for sparse tensors and PML Pytorch Metric Learning library 0.9.88~\cite{musgrave2020metric}.

\subsection{Results and discussion}
\label{sec:results}

\paragraph{Comparison with state-of-the-art.}

We compare performance our global descriptor with prior art: PointNetVLAD~\cite{angelina2018pointnetvlad}, PCAN~\cite{zhang2019pcan}, DAGC~\cite{sun2020dagc} and LPD-Net~\cite{liu2019lpd}. 
We also include DH3D~\cite{du2020dh3d} method in an evaluation.
DH3D is a recent 6DOF localization method, which includes a global point cloud descriptor computation as a part of the pose estimation pipeline.
%LPD-Net has state-of-the-art performance, but it must be noted, that it's is not fully learning-based method, as it heavily relies on hand crafted features.
%It must be noted, that all above methods are fully learning-based, with the exception of LPD-Net that heavily relies on hand crafted features.

Tab.~\ref{jk:tab:results_baseline} compares performance of our MinkLoc3D with state-of-the-art methods trained on the Baseline Dataset. When evaluated on Oxford dataset, MinkLoc3D wins with AR@1\% 3.0 p.p. higher than the runner-up, LPD-Net. 
When evaluated on three in-house datasets, it performs slightly worse compared than LPD-net (1.0 and 0.6 p.p. worse for U.S. and B.D. sets respectively and 0.7 p.p. better for R.A.). 
It must be noted that Oxford dataset and three in-house datasets were acquired using LiDARs with different characteristics. Even so, our method yields comparable results to LPD-Net which relies on hand-crafted features.
MinkLoc3D discriminability and generalization capability is significantly higher than all other fully learning based methods. 

\begin{table}[htbp]
\begin{center}
\begin{tabular}{l@{\quad}r@{\quad}r@{\quad}r@{\quad}r}
\hline
& \begin{tabular}{@{}c@{}}Oxford \end{tabular}
& \begin{tabular}{@{}c@{}}U.S. \end{tabular}
& \begin{tabular}{@{}c@{}}R.A. \end{tabular}
& \begin{tabular}{@{}c@{}}B.D. \end{tabular} 
\\
[2pt]
\hline
PointNetVLAD~\cite{angelina2018pointnetvlad} & 80.3 & 72.6 &  60.3 &  65.3 \\
PCAN~\cite{zhang2019pcan} & 83.8 & 79.1 & 71.2 &  66.8  \\
DH3D-4096~\cite{du2020dh3d} & 84.3 & - & - & - \\
DAGC~\cite{sun2020dagc} & 87.5 & 83.5 & 75.7 & 71.2  \\
LPD-Net~\cite{liu2019lpd} & 94.9 & \textbf{96.0} & 90.5 & \textbf{89.1} \\
%minkfpngem6.txt, exp_8_3_1.txt, model_MinkFPNGeM_20200927_1633, 97.48, 93.36, 94.14, 85.88, 89.45, 80.31, 86.42, 80.47, 91.87, 85.00 LATEST MODEL (20190928)
%MinkLoc3D (our) & \textbf{97.5} & 94.1 & 89.5 & 86.4 \\
%wacv_minkfpngem_2.txt, config_baseline13.txt, model_MinkFPNGeM_20201108_0233, 97.91, 93.76, 95.04, 86.01, 91.19, 81.11, 88.48, 82.66, 93.16, 85.88
MinkLoc3D (our) & \textbf{97.9} & 95.0 & \textbf{91.2} & 88.5 \\
[2pt]
\hline
\end{tabular}
\end{center}
\caption{Evaluation results (Average Recall at 1\%) of place recognition methods trained on the Baseline Dataset.}
\label{jk:tab:results_baseline}
\end{table}

\begin{figure*}[tb]
\centering
\subfloat[Oxford\label{1a}]{%
\includegraphics[height=5.5cm]{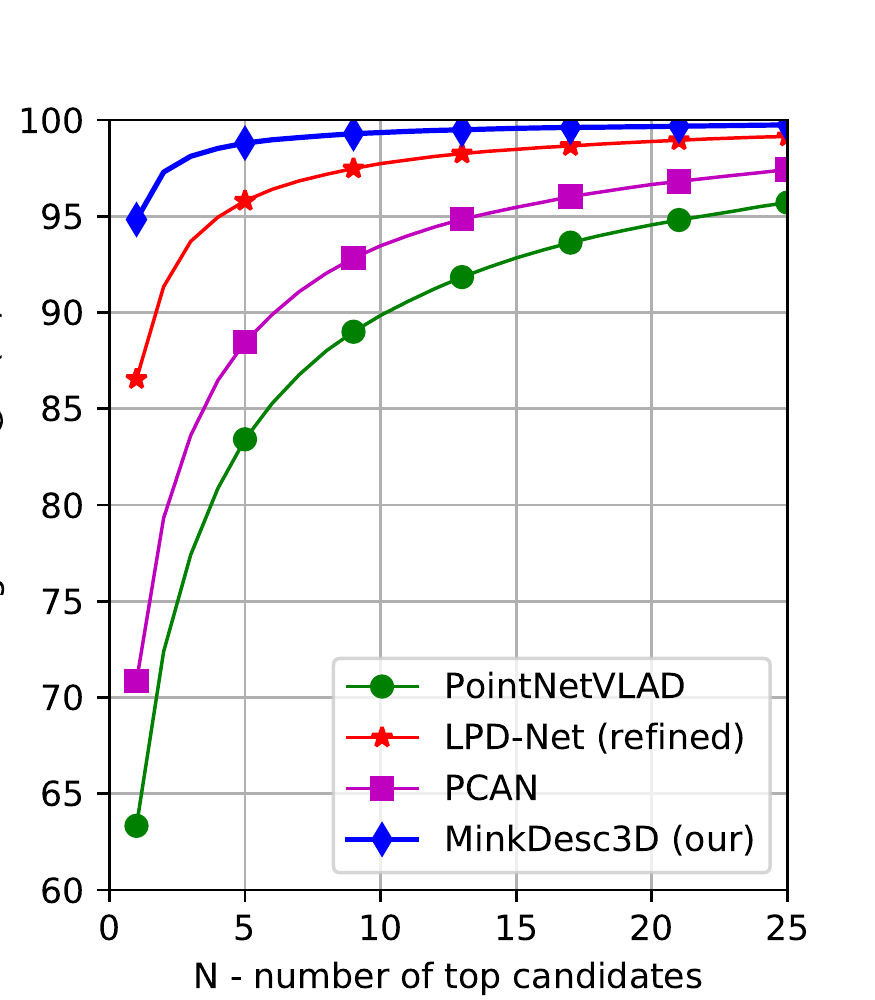}}
\hfill
\subfloat[U.S.\label{1b}]{%
\includegraphics[height=5.5cm]{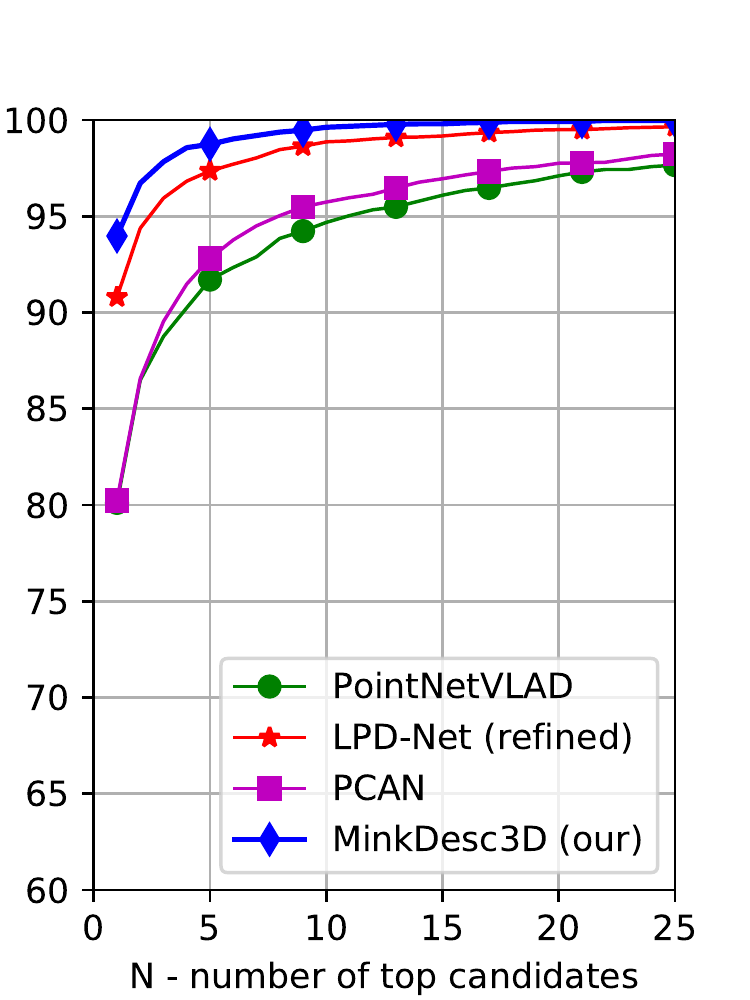}}
\hfill
\subfloat[R.A.\label{1c}]{%
\includegraphics[height=5.5cm]{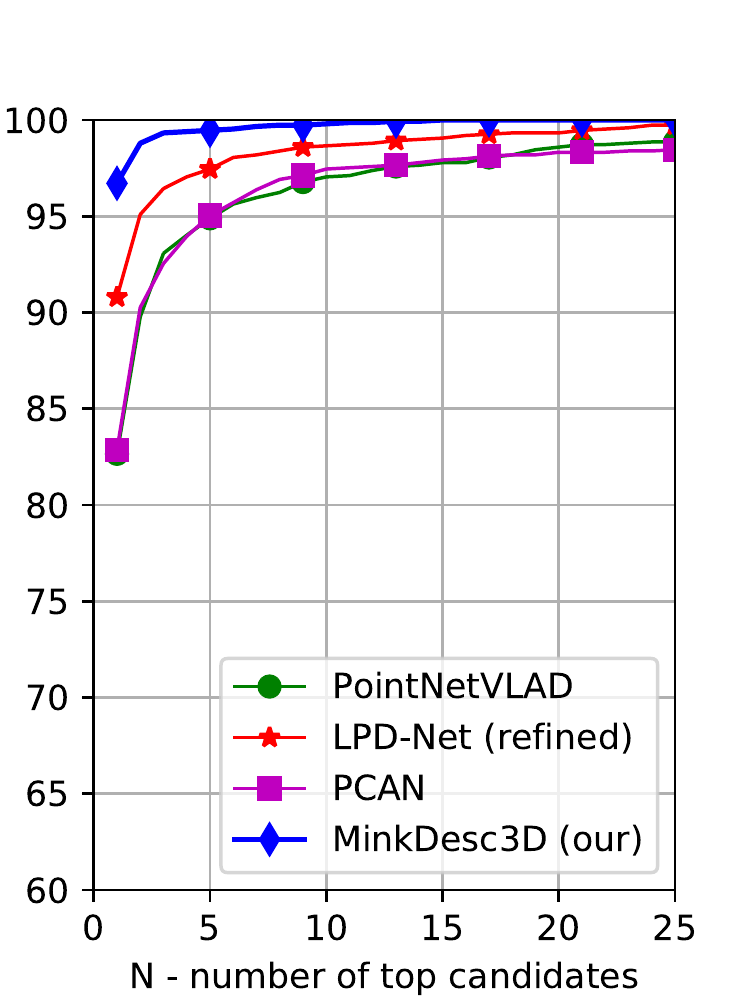}}
\hfill
\subfloat[B.D.\label{1d}]{%
\includegraphics[height=5.5cm]{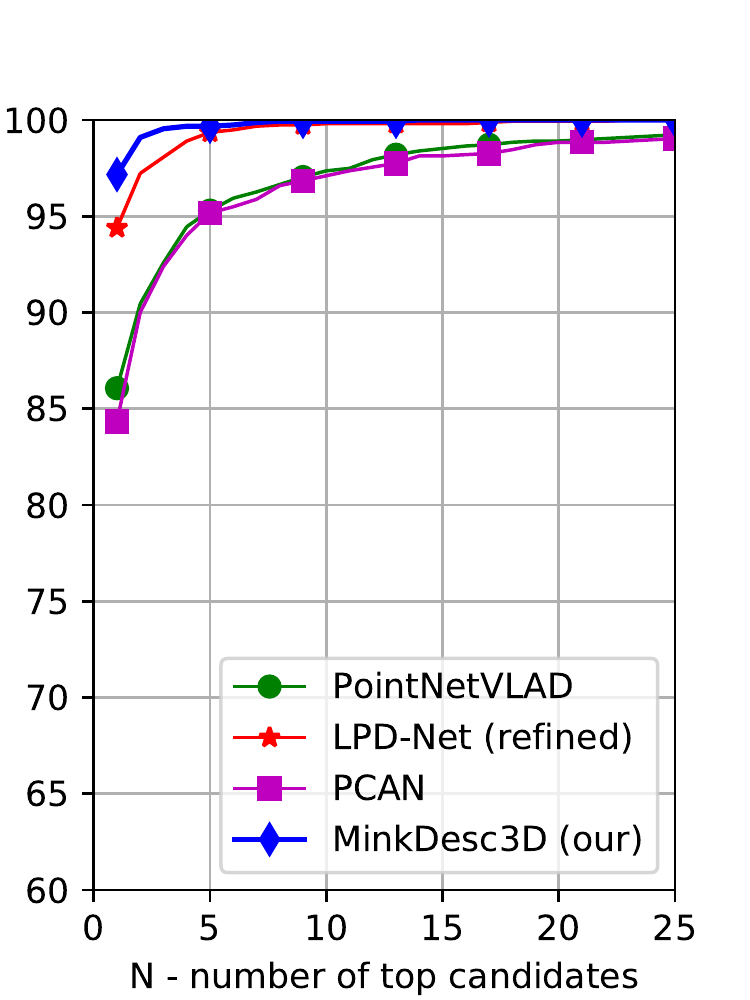}}
\caption{Average recall of place recognition methods trained on the Refined Dataset.}
\label{fig:plots_refined}
\end{figure*}

Tab.~\ref{jk:tab:results_refine} shows evaluation results of state-of-the-art methods trained on a larger and more diverse Refined Dataset. For PointNetVLAD and PCAN we run the evaluation using the trained models provided by authors. LPD-Net was trained from scratch and evaluated on Refined Dataset using the open source code.
Our MinkLoc3D is a clear winner. Compared to the state-of-the-art LPD-Net, the AR@1\% is higher between 0.8 p.p. and 2.9 p.p. on all evaluation subsets. 
The advantage over other methods is even higher, between 5-18 p.p.
Average Recall plots in Fig.~\ref{fig:plots_refined} show that our method outperforms previous methods on all evaluation subsets.

\begin{table*}[htbp]
\begin{center}
\begin{tabular}{l@{\quad}r@{\quad}r@{\quad}r@{\quad}r@{\quad}r@{\quad}r@{\quad}r@{\quad}r}
\hline
& \multicolumn{2}{c}{Oxford} & \multicolumn{2}{c}{U.S.}  & \multicolumn{2}{c}{R.A.} & \multicolumn{2}{c}{B.D.} \\
& \begin{tabular}{@{}c@{}}AR@1\% \end{tabular}
& \begin{tabular}{@{}c@{}}AR@1 \end{tabular}
& \begin{tabular}{@{}c@{}}AR@1\% \end{tabular}
& \begin{tabular}{@{}c@{}}AR@1 \end{tabular}
& \begin{tabular}{@{}c@{}}AR@1\% \end{tabular}
& \begin{tabular}{@{}c@{}}AR@1 \end{tabular}
& \begin{tabular}{@{}c@{}}AR@1\% \end{tabular}
& \begin{tabular}{@{}c@{}}AR@1 \end{tabular}
\\[2pt]
\hline
PointNetVLAD~\cite{angelina2018pointnetvlad} & 80.1 & 63.3 & 94.5 & 86.1 & 93.1 & 82.7 & 86.5 & 80.1 \\
PCAN~\cite{zhang2019pcan} & 86.4 & 70.7 & 94.1 & 83.7 & 92.3 & 82.3 & 87.0 & 80.3 \\
DAGC~\cite{sun2020dagc} & 87.8 & 71.5 & 94.3 & 86.3 & 93.4 & 82.8 & 88.5 & 81.3 \\
LPD-Net~\cite{liu2019lpd} & 94.9 & 86.6 & 98.9 & 94.4 & 96.4 & 90.8 & 94.4 & 90.8 \\
%minkfpngem6.txt, exp_101.txt, model_MinkFPNGeM_20200927_1350, 97.75, 93.41, 99.42, 96.26, 98.72, 95.84, 96.65, 93.60, 98.14, 94.78
%MinkLoc3D (our) & \textbf{97.8} & \textbf{93.4} & \textbf{99.4} & \textbf{96.3} & \textbf{98.7} & \textbf{95.8} & \textbf{96.7} & \textbf{93.6} \\[2pt]
%wacv_minkfpngem_2.txt, config_refined13.txt, model_MinkFPNGeM_20201108_0946, 98.50, 94.83, 99.68, 97.16, 99.33, 96.71, 96.72, 93.97, 98.56, 95.67
MinkLoc3D (our) & \textbf{98.5} & \textbf{94.8} & \textbf{99.7} & \textbf{97.2} & \textbf{99.3} & \textbf{96.7} & \textbf{96.7} & \textbf{94.0} \\[2pt]
\hline
\end{tabular}
\end{center}
\caption{Evaluation results (Average Recall at 1\% and at 1) of place recognition methods trained on the Refined Dataset.}
\label{jk:tab:results_refine}
\end{table*}

\captionsetup[subfigure]{labelformat=parens}

Tab.~\ref{jk:tab:resources} compares the number of trainable parameters and inference time (runtime per cloud).
Our MinkLoc3D method is significantly faster compared to LPD-Net. 
LPD-Net requires time consuming preprocessing of the input point cloud to compute 10 handcrafted features.
Even without including hand-crafted feature extraction time, LPD-Net has longer inference time compared to MinkLoc3D (26 vs 22 ms).
Our model is also much lighter compared to prior methods.
It has only 1.5 million trainable parameters, whereas other methods have an order of magnitude more.
This can be explained by the fact, that our method produces informative local features, that can be pooled using a simple Generalized-Mean pooling~\cite{miech17loupe} which has few learnable parameters.
Other methods use NetVLAD~\cite{arandjelovic2016netvlad} aggregation layer with millions of learnable parameters.

\begin{table}[htbp]
\begin{center}
\begin{tabular}{l@{\quad}r@{\quad}r@{\quad}r}
\hline
& \begin{tabular}{@{}c@{}}Parameters \end{tabular}
& \begin{tabular}{@{}c@{}}Runtime\\per cloud \end{tabular} 
\\
[2pt]
\hline
PointNetVLAD~\cite{angelina2018pointnetvlad} & 19.8M &  15 ms\\
PCAN~\cite{zhang2019pcan} & 20.4M & 55 ms \\
LPD-Net~\cite{liu2019lpd} & 19.8M &  26 ms \\
LPD-Net~\cite{liu2019lpd} with f.e. & 19.8M &  917 ms \\
MinkLoc3D (our) &  1.1M &  21 ms \\
[2pt]
\hline
\end{tabular}
\end{center}
\caption{Computation time required by different methods. \emph{LPD-Net with f.e.} includes hand-crafted features extraction time.}
\label{jk:tab:resources}
\end{table}

\begin{figure*}[htbp]
\centering
\setcounter{subfigure}{0}
\subfloat[query]{%
\includegraphics[width=2.7cm,height=2cm,trim={1.5cm 1cm 0.5cm 2cm},clip,cfbox=black 0.5pt 0.5pt]{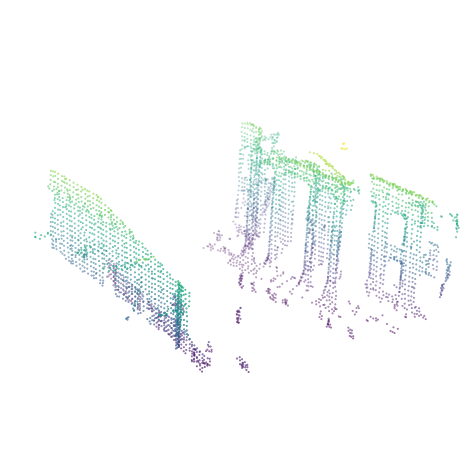}}
\hfill
\subfloat[dist=1.13 TP]{%
\includegraphics[width=2.7cm,height=2cm,trim={1.5cm 1cm 0.5cm 2cm},clip,cfbox=mygreen 0.5pt 0.5pt]{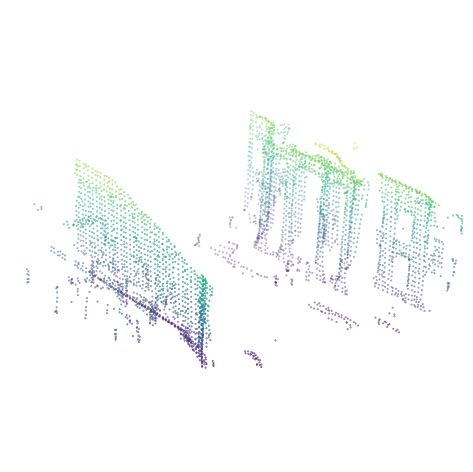}}
\hfill
\subfloat[dist=1.40 TP]{%
\includegraphics[width=2.7cm,height=2cm,trim={1.5cm 1cm 0.5cm 2cm},clip,cfbox=mygreen 0.5pt 0.5pt]{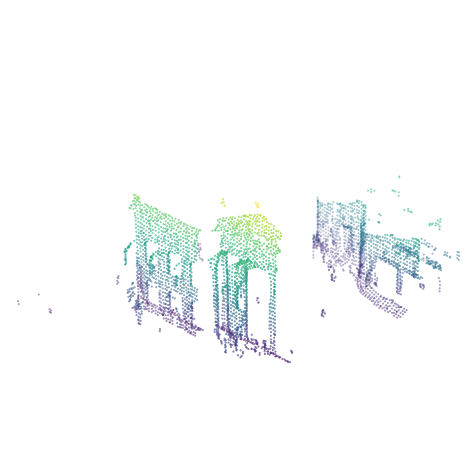}}
\hfill
\subfloat[dist=1.44 TP]{%
\includegraphics[width=2.7cm,height=2cm,trim={1.5cm 1cm 0.5cm 2cm},clip,cfbox=mygreen 0.5pt 0.5pt]{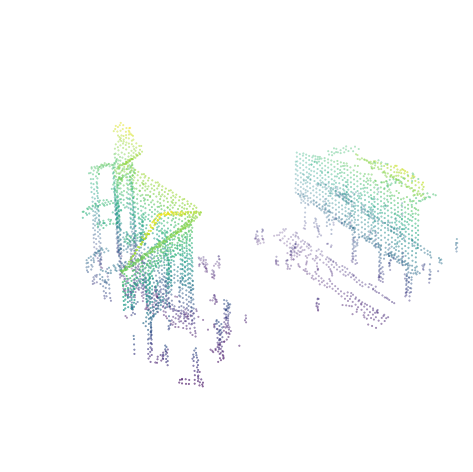}}
\hfill
\subfloat[dist=1.45 TP]{%
\includegraphics[width=2.7cm,height=2cm,trim={1.5cm 1cm 0.5cm 2cm},clip,cfbox=mygreen 0.5pt 0.5pt]{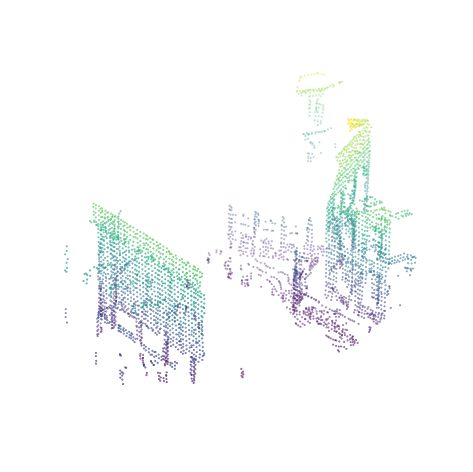}}
\hfill
\subfloat[dist=1.82 FP]{%
\includegraphics[width=2.7cm,height=2cm,trim={1.5cm 1cm 0.5cm 2cm},clip,cfbox=red 0.5pt 0.5pt]{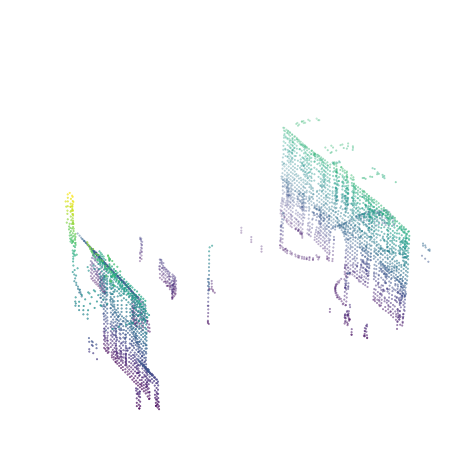}}

\caption{Nearest neighbours search results in Oxford evaluation subset. The leftmost column shows a query point cloud. Other columns show its five nearest neighbours. \emph{dist} is an Euclidean distance in the descriptor space. TP indicates true positive and FP false positive.}
\label{fig:search_results}
\end{figure*}

\begin{figure}[htbp]
\centering
\subfloat{%
\includegraphics[width=2.6cm,height=2cm,trim={1.5cm 1cm 0.5cm 2cm},clip,cfbox=black 0.5pt 0.5pt]{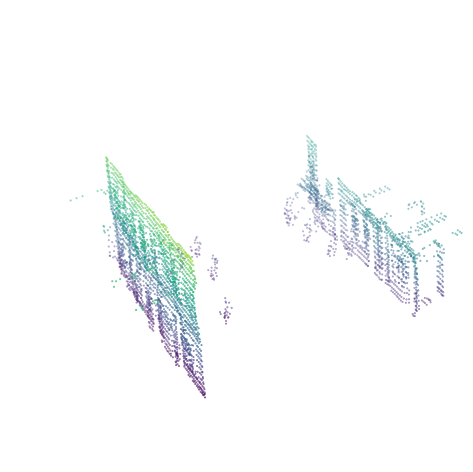}}
\hfill
\subfloat{%
\includegraphics[width=2.6cm,height=2cm,trim={1.5cm 1cm 0.5cm 2cm},clip,cfbox=red 0.5pt 0.5pt]{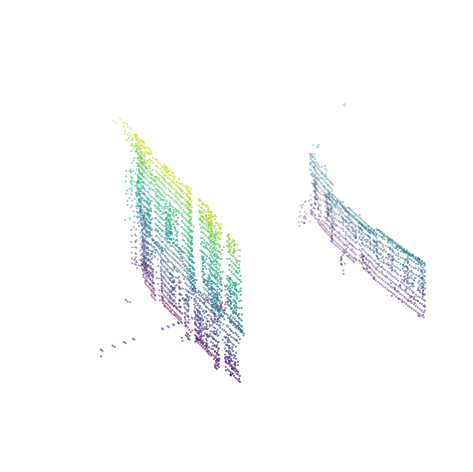}}
\hfill
\subfloat{%
\includegraphics[width=2.6cm,height=2cm,trim={1.5cm 1cm 0.5cm 2cm},clip,cfbox=mygreen 0.5pt 0.5pt]{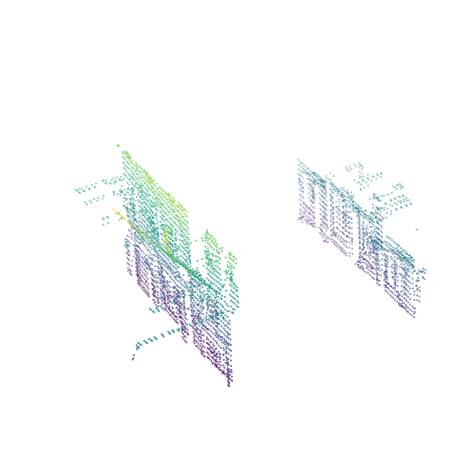}}
\\[-2ex]
\setcounter{subfigure}{0}
\subfloat[\label{11a}]{%
\includegraphics[width=2.6cm,height=2cm,trim={1.5cm 1cm 2cm 2cm},clip,cfbox=black 0.5pt 0.5pt]{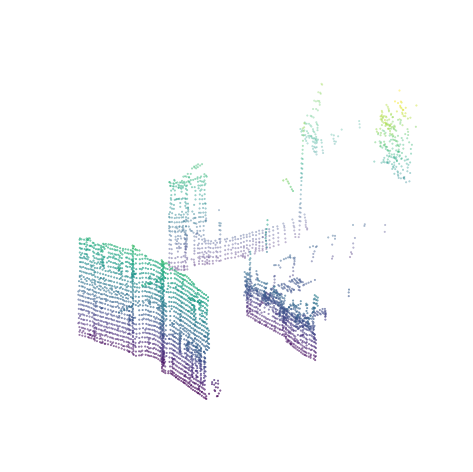}}
\hfill
\subfloat[\label{11b}]{%
\includegraphics[width=2.6cm,height=2cm,trim={1.5cm 1cm 2cm 2cm},clip,cfbox=red 0.5pt 0.5pt]{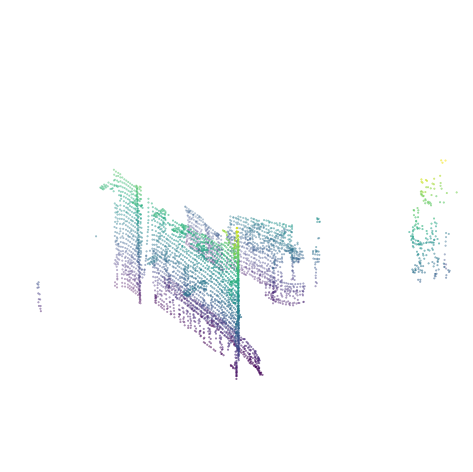}}
\hfill
\subfloat[\label{11c}]{%
 \includegraphics[width=2.6cm,height=2cm,trim={1.5cm 1cm 2cm 2},clip,cfbox=mygreen 0.5pt 0.5pt]{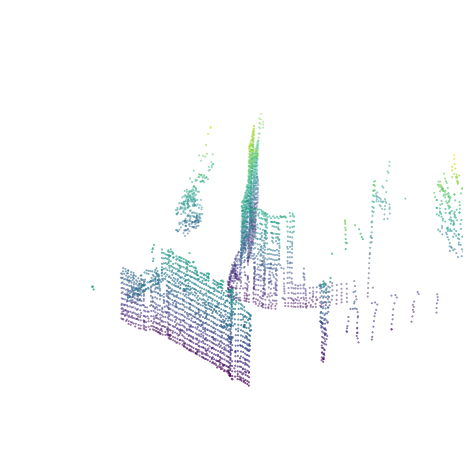}}
\caption{Failure cases. Examples of unsuccessful retrieval results using our network. (a) is the query point cloud, (b) incorrect match to the query and (c) the closest true match. }
\label{fig:failure_cases}
\end{figure}

Figure~\ref{fig:search_results} visualizes nearest neighbour search results using our MinkLoc3D descriptor in Oxford evaluation subset. 
The leftmost column shows a query point cloud and other columns show its five nearest neighbours. 
Figure~\ref{fig:failure_cases} shows failure cases. More visualizations of nearest neighbour search results can be found in Supplementary Material.

\captionsetup[subfigure]{labelformat=empty}

\captionsetup[subfigure]{labelformat=parens}

\paragraph{Ablation study.}

In this section we investigate impact of the network design choices on the discriminativity and generalization capability of our method.
In all experiments, the network is trained using the Baseline Dataset and evaluated on Oxford and three in-house datasets (U.S., R.A. and B.D.).

Tab.~\ref{jk:tab:results_aggregation} shows the impact of a feature aggregation method on the performance of the global descriptor. 
The following methods are evaluated: global max pooling (MAC), Generalized-Mean (GeM) pooling~\cite{radenovic2018fine}, NetVLAD~\cite{arandjelovic2016netvlad} and NetVLAD with Context Gating~\cite{miech17loupe} (NetVLAD-CG). 
Surprisingly, a simple GeM layer with few learnable parameters produces the most discriminative global descriptors and has the best generalization capability.
More sophisticated methods, NetVLAD and NetVLAD with Context Gating, score similarly on Oxford dataset, but noticeably worse on in-house datasets.
This can be attributed to two factors. 
First, our training datasets have a moderate size. Using NetVLAD layer, with millions of trainable parameters, increases the risk of overfitting.
Second, our local feature extraction network works very well and produces informative features, that can be effectively pooled using a simple GeM layer to produce a discriminative global descriptor.

%More sophisticated feature aggregation methods, such as NetVLAD with optional Context Gating, did not improved the performance. Average Recall@1\% on Oxford dataset is comparable to the results achieved with simple GeM feature pooling. The results on in-house datasets are significantly worse. This can be explained by the fact, that NetVLAD layer adds a large number of trainable parameters, 4.2M to the 1.5M parameters in local feature extraction block. This makes the network more prone to overfitting, hence worse generalization to in-house datasets. Adding the final CG layer add another 

%This can be explained by the fact, that our method produces informative local features, that can be pooled using a simple Generalized-Mean pooling~\cite{miech17loupe} to produce a discriminative global descriptor.
%Other methods use NetVLAD feature aggregation layer, which usually operates on 1024 dimensional local features and has millions of learnable parameters.

\begin{table}[htbp]
\begin{center}
\begin{tabular}{c@{\;}l@{\quad}r@{\quad}r}
\hline
& \begin{tabular}{@{}c@{}}Architecture\end{tabular}
& \begin{tabular}{@{}c@{}}Oxford\\AR@1\%\end{tabular}
& \begin{tabular}{@{}c@{}}In-house\\AR@1\%\end{tabular}
\\
[2pt]
\hline
%wacv_minkfpn_max.txt, config_baseline13.txt, model_MinkFPN_Max_20201108_1944, 97.29, 91.81, 94.72, 86.72, 89.85, 80.05, 87.78, 80.64, 92.41, 84.81 
& MinkFPN+MAC &  97.3 & 92.4 \\
%wacv_minkfpngem_2.txt, config_baseline13.txt, model_MinkFPNGeM_20201108_0233, 97.91, 93.76, 95.04, 86.01, 91.19, 81.11, 88.48, 82.66, 93.16, 85.88
* & MinkFPN+GeM & \textbf{97.9} & \textbf{93.2} \\
%wacv_minkfpn_netvlad.txt, config_baseline13.txt, model_MinkFPN_NetVlad_20201109_0500, 96.98, 90.56, 94.58, 82.73, 87.29, 75.26, 85.41, 78.32, 91.07, 81.72
& MinkFPN+NetVLAD & 97.0 & 91.1 \\
%wacv_minkfpn_netvlad_cg.txt, config_baseline13.txt, model_MinkFPN_NetVlad_CG_20201109_0030, 97.18, 91.34, 88.09, 72.36, 76.27, 62.03, 77.31, 67.53, 84.71, 73.31
& MinkFPN+NetVLAD-CG &  97.2 & 84.7 \\
[2pt]
\hline
\end{tabular}
\end{center}
\caption{Impact of a feature aggregation method on the discriminability of the global descriptor. The network is trained on the Baseline Dataset. * indicates MinkLoc3D architecture.}
\label{jk:tab:results_aggregation}
\end{table}

Tab.~\ref{jk:tab:desc_size} shows impact of a descriptor size on the discriminability of the global descriptor.
The number of channels in lateral connections (1x1Conv2, 1x1Conv3 blocks) and in a transposed convolution TConv3 block is set to the same value as the dimensionality of the final descriptor.  Parameters of bottom-up convolutional blocks remain unchanged.
The network performance is relatively similar with larger descriptor sizes (between 64 and 512) with AR@1\% between 97.3 and 98.0\% on Oxford dataset and between 90.3 and 93.2\% on in-house datasets. The performance deteriorates, when the descriptor size falls to 32.

\begin{table}[htbp]
\begin{center}
\begin{tabular}{cc@{\quad}r@{\quad}r@{\quad}r}
\hline
& \begin{tabular}{@{}c@{}}Descriptor\\size\end{tabular} & \begin{tabular}{@{}c@{}}Oxford\\AR@1\% \end{tabular}
& \begin{tabular}{@{}c@{}}In-house\\AR@1\% \end{tabular}
\\
[2pt]
\hline
%wacv_minkfpngem_2_out512.txt, config_baseline13.txt, model_MinkFPNGeM_20201108_2041, 97.99, 93.74, 95.36, 85.63, 90.19, 80.65, 86.65, 81.45, 92.55, 85.37
& 512 &  \textbf{98.0} & 92.6 \\
%256 & 97.5 & 90.0 \\
%wacv_minkfpngem_2.txt, config_baseline13.txt, model_MinkFPNGeM_20201108_0233, 97.91, 93.76, 95.04, 86.01, 91.19, 81.11, 88.48, 82.66, 93.16, 85.88
* & 256 & 97.9 & \textbf{93.2} \\
%minkfpngem14.txt, exp_8_3_1.txt, model_MinkFPNGeM_20200928_1053,97.32, 92.30, 94.01, 83.88, 87.50, 76.75, 83.39, 76.08, 90.55, 82.25
%128 & 97.3 & 88.3 \\
%wacv_minkfpngem_2_out128.txt, config_baseline13.txt, model_MinkFPNGeM_20201108_1702, 97.49, 92.90, 93.56, 83.83, 88.58, 78.83, 86.24, 79.76, 91.47, 83.83
& 128 & 97.5 & 91.5 \\
%minkfpngem15.txt, exp_8_3_1.txt, model_MinkFPNGeM_20200928_1358, 96.80, 89.85, 91.24, 80.34, 85.15, 73.05, 80.82, 73.10, 88.50, 79.09
%64 & 96.8 & 85.7 \\
%wacv_minkfpngem_2_out64.txt, config_baseline13.txt, model_MinkFPNGeM_20201108_1318, 97.27, 92.00, 93.75, 81.64, 86.90, 74.86, 83.41, 76.38, 90.33, 81.22
& 64 & 97.3 & 90.3 \\
%minkfpngem16.txt, exp_8_3_1.txt, model_MinkFPNGeM_20200928_1701, 95.75, 86.56, 87.62, 71.90, 78.29, 63.57, 76.58, 67.55, 84.56, 72.39
%32 & 95.8 & 80.8 \\
%wacv_minkfpngem_2_out32.txt, config_baseline13.txt, model_MinkFPNGeM_20201108_0946, 95.83, 87.46, 90.33, 76.61, 78.75, 64.72, 80.54, 71.00, 86.36, 74.95
& 32 & 95.8 & 86.4 \\
[2pt]
\hline
\end{tabular}
\end{center}
\caption{Impact of a descriptor size on the discriminability of the global descriptor. The network is trained on the Baseline Dataset.}
\label{jk:tab:desc_size}
\end{table}

%Results of experimental evaluation of deeper and wider variants of MinkLoc3D network can be found in Supplementary Material.

\paragraph{Comparison with image based methods.}

In this paragraph we compare performance of our MinkLoc3D with state-of-the-art image-based place recognition and visual localization methods in challenging environmental conditions. 
The comparison is done using RobotCar Seasons~\cite{sattler2018benchmarking} dataset.
It contains outdoor images captured in the city of Oxford at various periods of a year in different atmospheric conditions, e.g. snow, rain, dawn or night.

\begin{table*}[htbp]
\begin{center}
\begin{tabular}{l@{\quad}|r@{\quad}r@{\quad}r@{\quad}r@{\quad}r@{\quad}r@{\quad}r@{\quad}|r@{\quad}r}
%\hline
&  \multicolumn{7}{c}{day conditions} & \multicolumn{2}{c}{night conditions} \\
&  \begin{tabular}{@{}c@{}}dawn\end{tabular}
& \begin{tabular}{@{}c@{}}dusk\end{tabular}
& \begin{tabular}{@{}c@{}}overcast\\summer\end{tabular}
& \begin{tabular}{@{}c@{}}overcast\\winter\end{tabular}
& \begin{tabular}{@{}c@{}}rain\end{tabular}
& \begin{tabular}{@{}c@{}}snow\end{tabular}
& \begin{tabular}{@{}c@{}}sun\end{tabular}
& \begin{tabular}{@{}c@{}}night\end{tabular}
& \begin{tabular}{@{}c@{}}night-rain\end{tabular}
\\[2pt]
\hline
DenseVLAD~\cite{torii201524} & 92.5 & 94.2 & 92.0 & 93.3 & 96.9 & 90.2 & 80.2 & 19.9 & 25.5 \\
NetVLAD~\cite{arandjelovic2016netvlad} & 82.6 & 92.9 & 95.2 & 92.6 & 96.0 & 91.8 & 86.7 & 15.5 & 16.4 \\
\hline
NetVLAD+SP~\cite{sarlin2019coarse} &  90.3 &  96.7 & 98.1 & 96.2 & 97.6 & 95.9 & 94.1 & 35.4 & 33.4\\
DenseVLAD+D2-Net~\cite{dusmanu2019d2} & 94.4 & 95.9 & 98.3 & 96.2 & 96.9 & 94.9 & 91.1 & 53.9 & 56.1 \\
NetVLAD+SP+SG~\cite{sarlin2020superglue} & 97.3 & 97.2 & 99.8 & 96.7 & 98.1 & 97.8 & 96.1 & 91.9 & 92.0 \\
\hline
LPD-Net~\cite{liu2019lpd}  & 79.7 & 79.9 & 79.7 & 73.8 & - & - & 82.3 & 77.3 & 32.8 \\
%Method	night all	night rain	overcast winter	sun	rain	snow	dawn	dusk	night	day all	overcast summer
%MinkLoc3D	1.7 / 8.9 / 72.0	0.2 / 4.8 / 58.0	2.1 / 11.5 / 83.1	4.6 / 12.6 / 87.4	1.9 / 10.5 / 66.3	3.7 / 13.5 / 86.3	3.5 / 18.0 / 89.2	4.1 / 14.7 / 88.3	3.2 / 13.0 / 86.1	3.1 / 13.0 / 84.6	1.9 / 9.7 / 90.3
MinkLoc3D (our) & 89.2 & 88.3 & 90.3 & 83.1 & 66.3 & 86.3 & 87.4 & 86.1 & 58.0 \\ [2pt]
\hline
\end{tabular}
\end{center}
\caption{Comparisons with 6DoF visual localization methods on RobotCar Seasons dataset. We report percentage of queries correctly localized within 5 meter and 10$^{\circ}$ threshold.
Five top rows show performance of image-based methods and two bottom rows LiDAR scan-based methods.}
\label{jk:tab:results-seasons}
\end{table*}

We compare MinkLoc3D performance against place recognition methods based on a global image descriptor: 
DenseVLAD~\cite{torii201524} and NetVLAD~\cite{arandjelovic2016netvlad};
and against full 6DoF (6 degree-of-freedom) relocalization methods: 
NetVLAD+SP~\cite{sarlin2019coarse}, 
DenseVLAD+D2-Net~\cite{dusmanu2019d2} and
NetVLAD+SP+SG~\cite{sarlin2020superglue}.
For each image in RobotCar Seasons dataset, we find LiDAR readings with corresponding timestamps in the in original RobotCar dataset~\cite{RobotCarDatasetIJRR} and construct the point cloud. 
Then, we use MinkLoc3D network to compute a global point cloud descriptors and link this descriptor with a corresponding image.
To approximate a 6DoF pose of a query image, we search for a database image which descriptor (computed from its corresponding point cloud), is closest to the descriptor of a query image (computed from a corresponding point cloud). Then, we return the known pose of a database image as an approximation of the query image pose.
Retrieved poses are evaluated using the online evaluation service at \emph{Long-term visual localization} site.\footnote{https://www.visuallocalization.net/}

Results are shown in Tab.~\ref{jk:tab:results-seasons}. 
In day conditions image-based methods perform generally better, with up to 10 p.p. more correctly localized queries. However, both LiDAR based methods (ours and LPD-Net) operate on relative small, downsampled point clouds with 4096 points. Even with this small number of points both methods perform reasonably well.
Also both LiDAR based methods approximate 6DoF pose by taking the pose of the closest nearest neighbour found.
Full 6DoF localization methods, NetVLAD+SP, DenseVLAD+D2 Net and NetVLAD+SP+SG, employ much more sophisticated approach, where candidate matches found using a global descriptor are filtered by matching local features with geometric consistency criteria.
In night conditions point cloud-based method show their potential. 
Our method surpass all image-based methods by a large margin with the exception of the latest NetVLAD+SP+SG, which has slightly higher performance.
%It can be noticed, that rain adversely affects performance of both point cloud-based localization methods (ours and LPD-Net\cite{liu2019lpd}). 

\section{Conclusion}

In this paper we present MinkLoc3D, a novel 3D point cloud descriptor, based on a sparse voxelized point cloud representation and 3D FPN~\cite{lin2017feature} architecture. 
Extensive experimental evaluation proves that it outperforms prior cloud-based place recognition methods.
The success of our method can be attributed to two factors.
First, sparse convolutional architecture can produces informative local features, that can be used to construct a discriminative global point cloud descriptor.
%There's no need to use additional processing blocks, such as graph neural networks for multi-scale local feature aggregation, as in competitive methods.
Second, improvements in the training process allows efficient and effective training with larger batch size, which positively  affects discriminability and generalization capability of the resultant descriptor.
The natural next step is to enhance the proposed method to build full 6DoF localization solution.
%MinkLoc3D can be used to efficiently retrieve a few candidate matches. Similar architecture, based on sparse 3D convolutions, can be used to extract local feature that will be matched between candidate point clouds and the query cloud.

It should be also noted, that achieved results (AR@1\% between 96.7\% and 99.4\% when trained on Refined Dataset) show that standard benchmarks used to train and evaluate point cloud-based place recognition methods are close to being saturated and there's a little room for improvement. 
Larger and more diverse datasets would be needed to instigate further progress.

\paragraph{Acknowledgements}
The project was funded by POB Research Centre for Artificial Intelligence and Robotics of Warsaw University of Technology within the Excellence Initiative Program - Research University (ID-UB).

{\small
\bibliographystyle{ieee_fullname}
\bibliography{jk-bib}
}

\end{document}